\documentclass[a4paper,conference]{IEEEtran}

\usepackage{lscape}
\usepackage[table,xcdraw]{xcolor}
\usepackage{graphicx}

\begin{document}
%
\title{GGViT:Multistream Vision Transformer Network in Face2Face Facial Reenactment  Detection}


\author{
\IEEEauthorblockN{Haotian Wu$^{1,2}$,Peipei Wang$^{3}$,Xin Wang$^{1*}$,Ji Xiang$^{1}$,Rui Gong$^{1}$}
\IEEEauthorblockA{$^1$Institute of Information Engineering,Chinese Academy of Sciences, Beijing, China}
\IEEEauthorblockA{$^2$School of Cyber Security, University of Chinese Academy of Sciences, Beijing, China}
\IEEEauthorblockA{$^3$National Computer Network Emergency Response Technical Team\\Coordination Center of China (CNCERT/CC), Beijing, China}
\IEEEauthorblockA{Email:\{wuhaotian\}@iie.ac.cn, \{wangpeipei\}@cert.org.cn, \{wangxin,xiangji,gongrui\}@iie.ac.cn}
}


%


\maketitle

\begin{abstract}
Detecting  manipulated facial images and videos on social networks has been an urgent problem to be solved. The compression of videos on social media has destroyed some pixel details that could be used to detect forgeries. Hence, it is crucial to detect manipulated faces in videos of different quality. We propose a new multi-stream network architecture named GGViT, which utilizes global information to improve the generalization of the model. The embedding of the whole face extracted by ViT will guide each stream network. Through a large number of experiments, we have proved that our proposed model achieves state-of-the-art classification accuracy on FF++ dataset, and  has been greatly improved on scenarios of different compression rates. The accuracy of Raw/C23, Raw/C40 and C23/C40 was increased by 24.34\%, 15.08\% and 10.14\% respectively.
\end{abstract}


%
\IEEEpeerreviewmaketitle

\section{Introduction}
\begin{figure}[t]
	\begin{center}
		\includegraphics[width=0.9\linewidth]{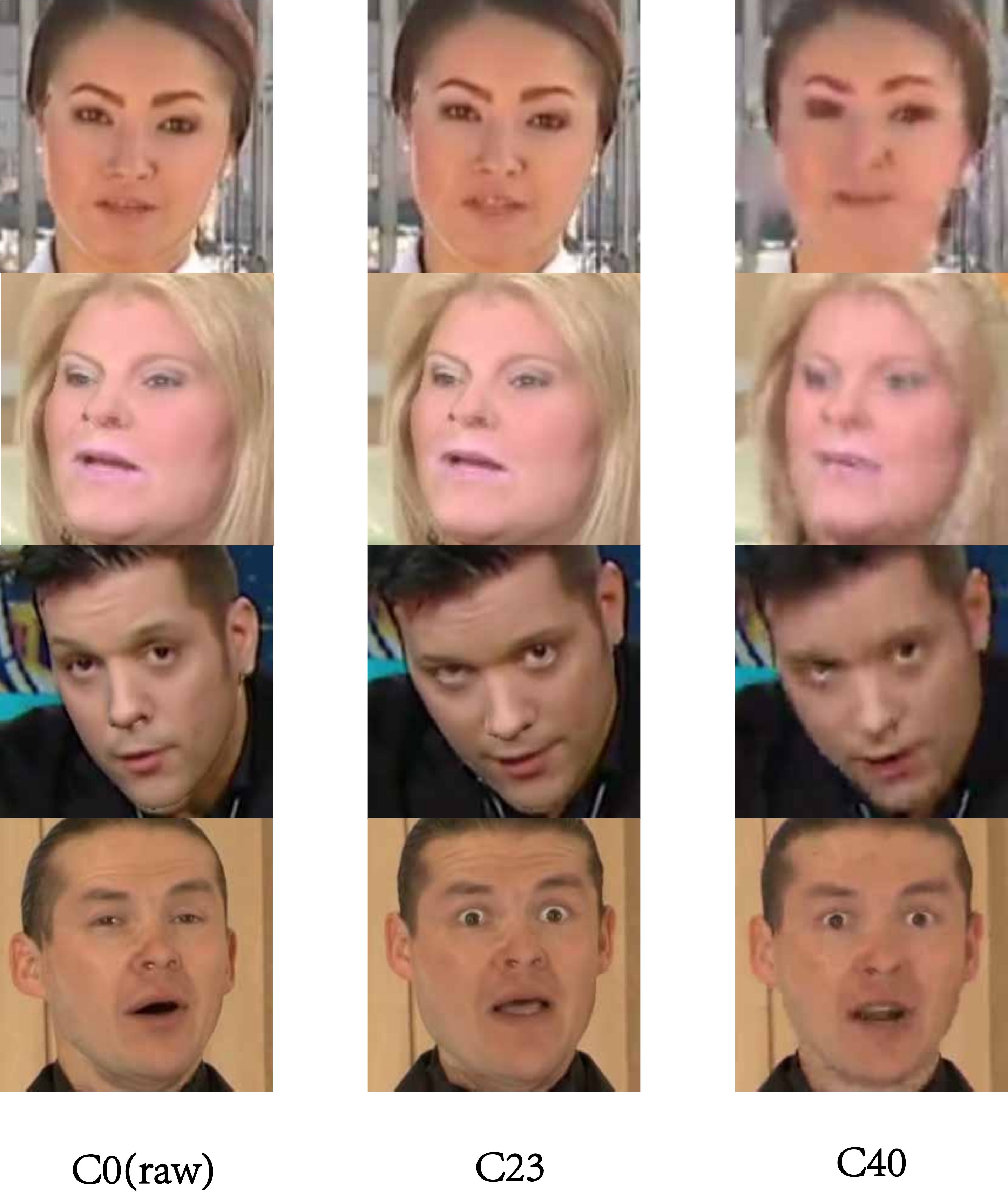}
	\end{center}
	\caption{Compressed images will reduce the image quality, and forged traces are more difficult to be observed. The four rows of forged images show the traces of the edge of the cheek, the mouth, the chin and the triangular area of the mouth, which are more difficult to detect with the increase of the image compression rate. The compression parameters are 0 (no compression), 23 and 40, respectively}
	\label{fig:intro}
\end{figure}
A large number of videos and pictures are transmitted through social networks all the time. People don't have enough time to identify the authenticity of each picture and video in the face of a large amount of information. Facial reenactment is a common form of forged faces. In the past, this kind of forged videos usually used splicing or synthesis methods. These methods not only need a lot of time to make, but also can be easily recognized by people. The development of deep learning network reduces the manufacturing obstacles to this technology, a large number of forged videos can be produced and transmitted quickly. In order to solve this problem, Many previous studies\cite{chollet2017xception}\cite{kumar2020detecting} have designed different models to extract features from these videos for classification. On this basis, facial movement information in the videos was also taken into account in classification\cite{komulainen2012face}\cite{li2018ictu}. Although these methods achieve excellent classification results from the same video quality datasets, different video quality in the wild brings a great challenge to these methods.

In order to reduce the pressure in data storage and improve the speed of information transmission, social media usually compresses the transmitted videos to varying degrees. Compression will reduce the quality of pictures, so that the forged videos will lose some forgery details that can be easily distinguished. Compression also makes it more difficult for existing methods to classify these forged videos(Fig.\ref{fig:intro}). Although great progress has been made in the classification of the same video quality, identifying forged videos in different compression rates is still a very necessary measure to protect the information security of social media. 

In this work, we present a new multi-stream network architecture based on ViT(Vision Transformer), named GGViT(Global Guidance Vision Transformer). Our approach mainly addresses the challenge to classification accuracy of facial reenactment in different video quality environments. To tackle this challenge, inspired by  Kumar\cite{kumar2020detecting} and ViT(vision transformer\cite{dosovitskiy2020image}), we propose a multi-stream network. In this network, the face to be detected will be divided into multiple parts, and the same part of each face will be classified by a special stream network to obtain better classification results. The central idea of GGViT is to add the embedding of whole face image to the other parts of the face image. The global information about the whole face can guide the stream network responsible for the local face. Considering the problem of image quality, we designed an image quality block to extract the image quality information, and made different constraints on the final classification results according to the image quality to enhance the generalization ability of the GGViT. The prediction results of multi-stream networks will be reconstructed by fusion attention block of the GGViT. More attention will be paid to the prediction results of stream networks that can better discriminate in favor of the corresponding image quality.

We conducted a large number of experiments on the FF++\cite{roessler2019faceforensicspp} dataset, and the experimental results show that GGViT achieves state-of-the-art on FF++ dataset, and has a great improvement in scenarios of different compression rates.

In summary, we make three major contributions in this
paper:
\begin{itemize}
	\item  We use ViT models to design a multi-stream deep learning network to detect facial reenactment in videos.
 \item We propose a loss function that has excellent performance for the network of different compression rates.
\item A large number of experiments have been conducted on the proposed model, and our proposed method has reached state-of-the-art on mainstream datasets.
\end{itemize}
\section{Related Work}
In recent years, face forgery has received more attention due to its wide range of applications. Correspondingly, face forgery detection has also become a popular research field. In this section, we will briefly review the evolution of face reenactment technology and the progress of corresponding face reenactment detection methods.
\subsection{Face Reenactment Generation Techniques}
The method of face reenactment refers to the transfer of the source facial expression to the target face without changing the identity of the target. It can be roughly divided into a method using three-dimensional models and a method based on GAN\cite{goodfellow2014generative}. Suwajanakorn\cite{suwajanakorn2017synthesizing} produced photorealistic results by using audio to produce lip movements, combining proper 3-D pose with high quality lip textures and mouth shape. Volker\cite{blanz1999morphable} derived a morphable face model by transforming the shape and texture of the examples to a vector space representation. Face2Face\cite{thies2016face2face} effectively transferred the expressions of target face and source face through a transfer matrix, toke into account the details of mouth opening, and re-renders and syntheses the faces with changed expressions.

The GAN-based method requires a large number of paired images for training. Jin\cite{jin2017cyclegan} directly used CycleGAN\cite{zhu2017unpaired} to exchange expressions between faces of different identities, capture details of facial expressions and head poses to generate transformation videos of higher consistency and stability. ReenactGAN\cite{wu2018reenactgan} used the mapping of the underlying space to transfer facial movements and expressions from an arbitrary person's monocular video input to a target person's video in real-time.
\begin{figure}[t]
	\begin{center}
		\includegraphics[width=0.9\linewidth]{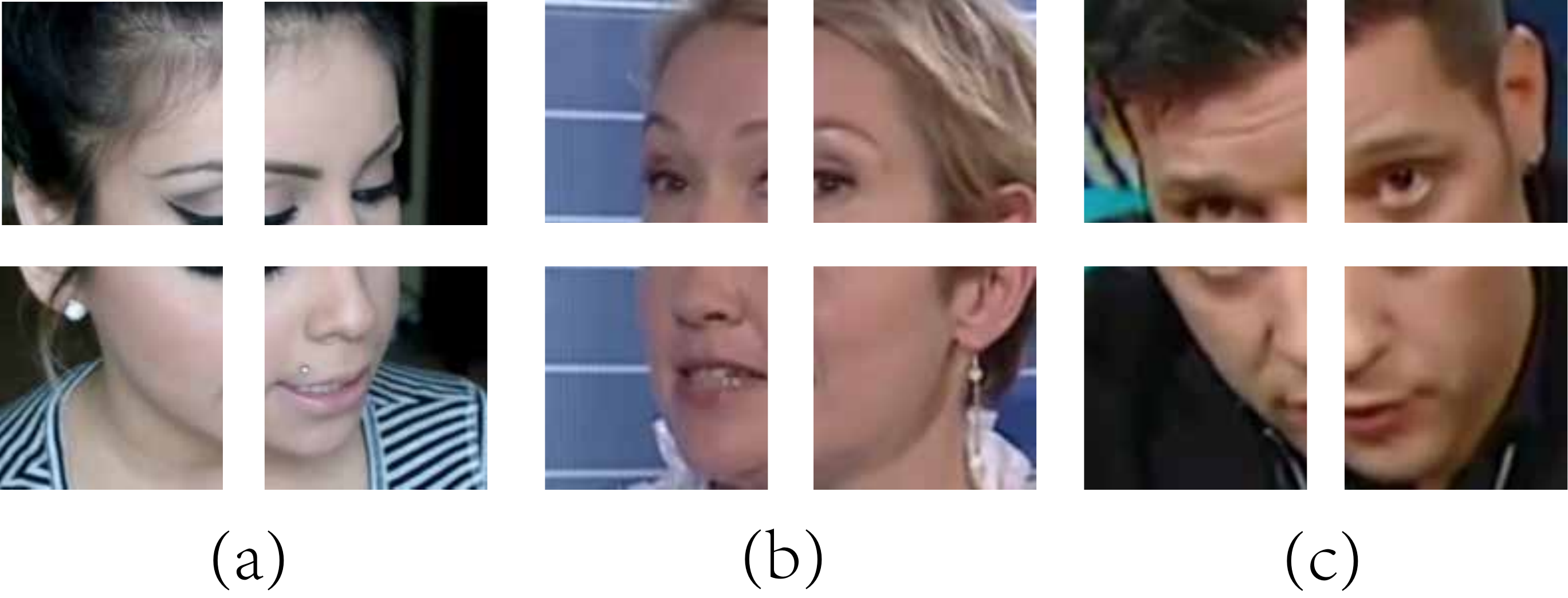}
	\end{center}
	\caption{In real application scenarios, the obtained faces are not aligned. After such a picture is divided into four parts, the network responsible for detecting the eye part can only get a picture of half an eye, while the nose part occasionally appears completely at the lower left of (a) and occasionally at the lower right of (b), which increases the detection difficulty of each stream network.}
	\label{fig:method}
\end{figure}
\begin{figure*}[t]
	\begin{center}
		\includegraphics[width=0.9\linewidth]{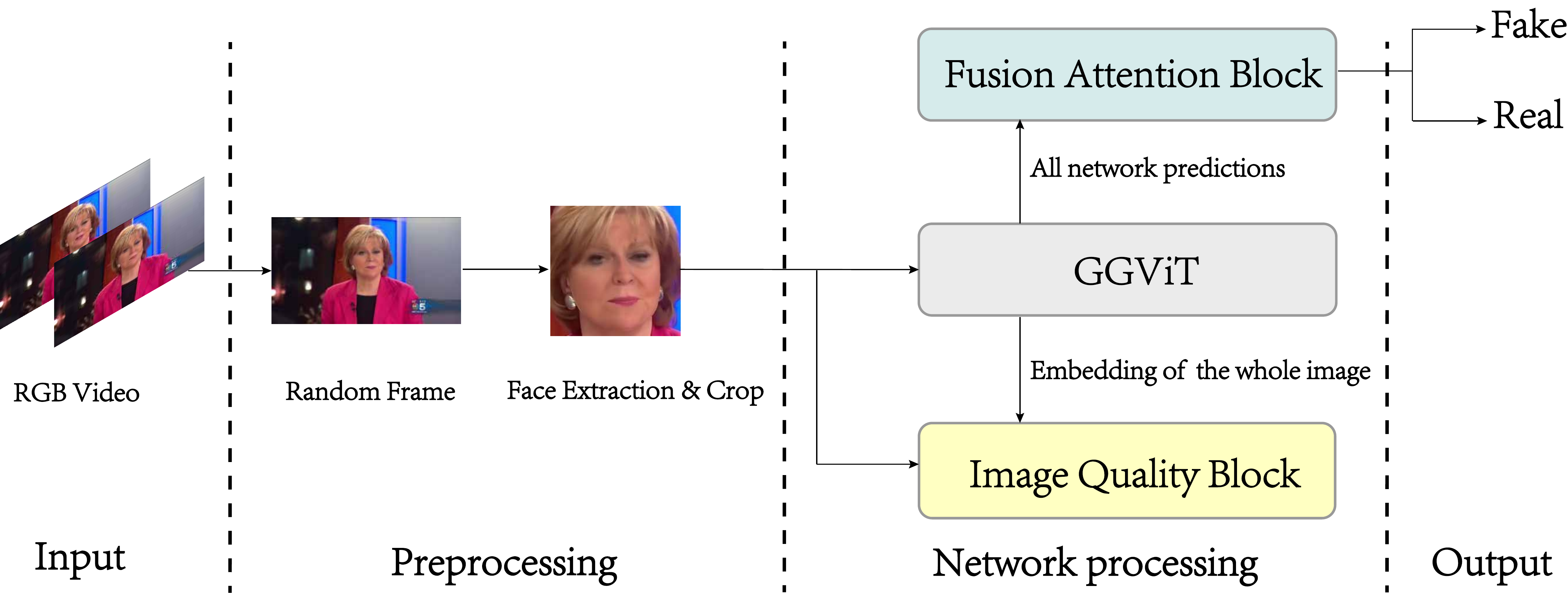}
	\end{center}
	\caption{The picture shows the process of inputting the video to be detected to outputting the result. Image quality block is used to calculate $L_{lmc}$. Fusion attention block integrates the output results of multi-stream network and outputs the final result}
	\label{fig:method1}
\end{figure*}

\subsection{Face Reenactment Detection Techniques}
The traditional face forgery detection technology obtains artificial features from face images for discrimination, uses LBP\cite{boulkenafet2015face}, SIFT\cite{patel2016secure} and other local descriptors to obtain image features, and then uses SVM\cite{joachims1998making} or other methods of discrimination. In the stage of producing a large number of forged pictures and videos, the deep neural network is mainly used to extract features and judge the traces generated by forged pictures. Matern\cite{matern2019exploiting} built on this by taking into account differences in physical features such as eyes and teeth. According to the characteristics of forged videos, dynamic information, such as dynamic texture\cite{komulainen2012face}, twitching\cite{siddiqui2016face} and blinking\cite{li2018ictu}, muscle\cite{agarwal2019protecting} and other dynamic elements are used to assist identification.
With the development of deep learning, multi-stream network was gradually used to distinguish forged images considering different reference factors of the network. Zhou\cite{zhou2017two} took local noise residuals and camera characteristics as a second stream. Atuom\cite{atoum2017face} proposed a novel two-stream CNN-based approach, which takes the local features and global depth of images as the input of two networks. Masi\cite{masi2020two} used two-stream network to suppress the influence of simple facial information on network output. Kumar\cite{kumar2020detecting} had achieved good results by dividing faces into multiple parts and using a multi-stream network to focus on local forgeries.

\section{Methods}

In this section, we will first state our design motivation and briefly introduce our framework. As mentioned earlier, the differences between real and reenactment images are subtle and often localized. Moreover, the operation method of face reenactment usually leaves traces in the canthus, chin, cheek and other parts. Therefore, it is very beneficial to divide the face into different partial pictures and train specialized networks to find the corresponding reenactment feature of each similar facial part.

However, the face videos obtained from social networks are likely not aligned faces in actual scenes. Therefore, if the face is partially converted into four local images, the face parts contained in the local images are likely to be misplaced or missing( Fig.\ref{fig:method} ). Local network judgment is likely to be wrong due to the lack of global information. Combining global information with local images can help local networks to make better judgments.

Under different video compression rates, the amount of forged details contained in different image quality is different. Image quality can be regarded as the confidence of features extracted from the network, and the network should have corresponded discrimination standards for images of different quality.

According to the above observation, for the face image extracted from the video. We designed such a multi-stream network for the identification of face reenactment videos. As can be seen in Fig.\ref{fig:method1} :
\begin{itemize}
\item We used five ViT-B/16 models to construct what we call a GGViT network. One of the ViT-B/16 models was used to extract the information of the whole face, and the extracted features were used to guide the other four networks to focus on four equal square images divided by the whole picture respectively.
\item We designed an image quality block, which carries out different constraints on classification by detecting image quality information, so that the model can obtain better generalization ability.
\item For the output results of multi-stream networks, the fusion attention block was designed to integrate the results of different stream networks to make the final judgment.
\end{itemize}

\subsection{Preprocessing}
For the frame images of human faces randomly extracted from input videos, Retinaface\cite{deng2020retinaface} was used to detect human faces. For the case of multiple faces in one image, the center point of the detected face box is compared with that of the mask to determine the detected face position. The detected face box was first enlarged into a square with the length of the box as its side length, and then the square box were enlarged on a ratio of 1.1. Finally, the whole face image will be resized to 224 x 224 to keep the same size, and each whole face image will be equally divided into four square images containing only parts of the face, and the resulting partial face images will also be resized to 224 x 224. All five images will be used in next stage.
\begin{figure*}[t]
	\begin{center}
		\includegraphics[width=0.9\linewidth]{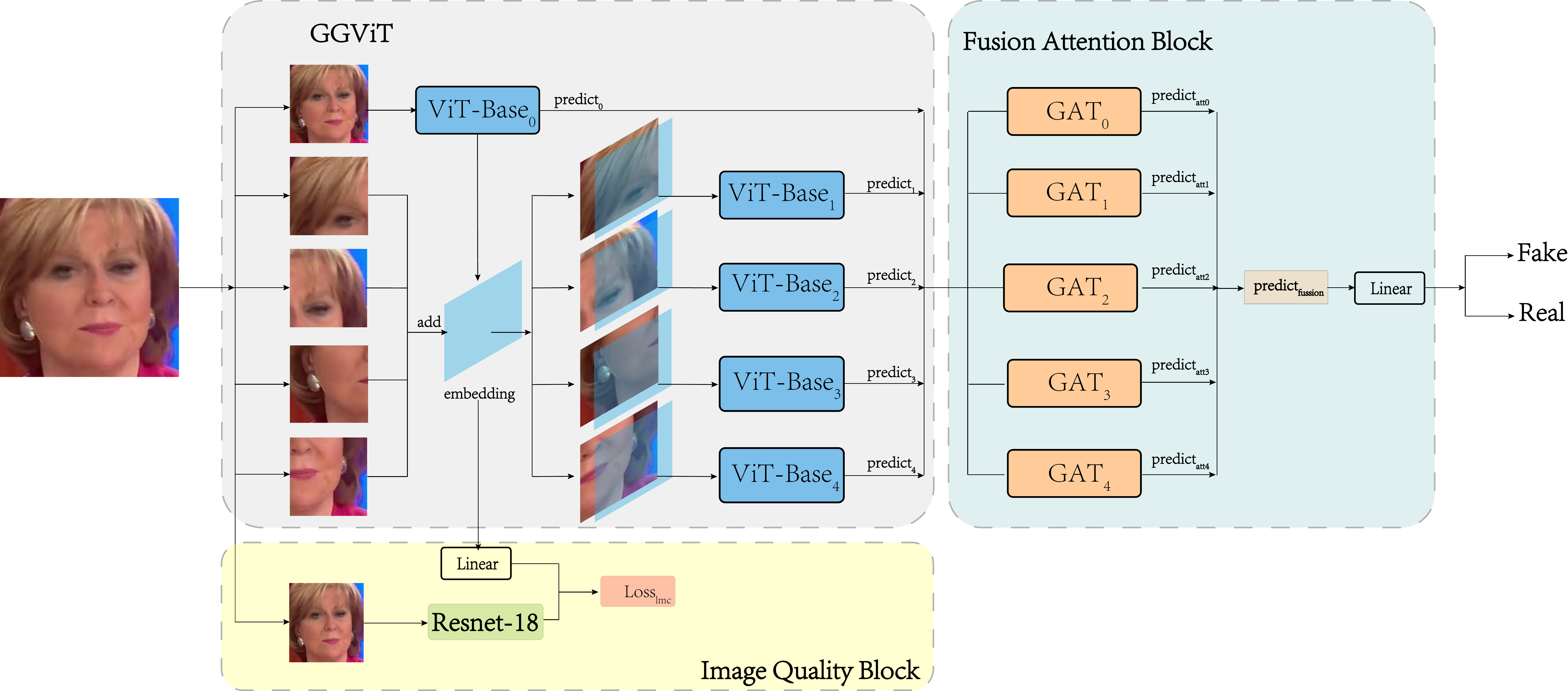}
	\end{center}
	\caption{All images will be processed to the size of 3x224x224, and the embedding of the whole face which from the ViT-$Base_0$ will be transformed to 3x224x224 too. The GGvit will add the embedding of the whole image to the other four local images before they are sent to the corresponding ViT network. Each stream network outputs its own prediction result. Fusion attention block makes a new prediction based on each prediction result with the other prediction results.}
	\label{fig:method2}
\end{figure*}
\subsection{Network processing}
The overall detection model is divided into three parts. Firstly, all five images are sent to the multi-stream network to detect whether they are forged respectively. The shunting network that detects the whole face also  provides the whole image's embedding to the image quality block to calculate the corresponding quality loss and constrain the network training. Fusion attention block integrates the output of the multi-stream network to produce the final result.
\subsubsection{GGViT}
GGViT consists of five VIT-B/16 models(Fig.\ref{fig:method2}), $ViT_0$ is used to detect the whole face, and the other four ViTs are used to detect the four parts of the face image. The size of the input image is 224 x 224, and $ViT_0$ is used to provide a 768-dimension embeddding of the whole image, which is converted into a 3 x 16 x 16 tensor. Then repeat it to get the tensor of 3 x 224 x 224, and then add the tensor to the other four partial images of faces. The stacked images are input to their respective ViT networks to get their respective predict. The output of $ViT_0$ plays a guiding role in local face image recognition, which is proved by our experiments.
\subsubsection{Image Quality Block}
In this module,  we used the FF++ verification set of images with different compression rates to train a classifier in advance. The classifier trains 20 epochs of ResNet-18\cite{he2016deep} by using pre-training weights to classify images with different compression rates. The output results can reflect the compression quality of images. The 768-dimension embedding of the whole face image obtained by ViT0 gets a 511-dimension tensor through a full connection layer, and the evaluation result of image quality constitutes a 512-dimension tensor with the 511-dimension vector for the subsequent loss calculation.
\subsubsection{Fusion Attention Block}
For the integration of the discriminant results of the five ViTs' output, five graph attention networks\cite{velivckovic2017graph} are used. The output results of the five ViTs are respectively regarded as the main node, and the other four results are regarded as neighbor nodes. Each graph attention network outputs a new predict that takes the other four classification results into account. Five 1x2 classification results are cat into a 1x10 fusion result, and then through one fully connection layer, the final classification result of the video can be obtained.

\subsection{Loss Function}
The loss function as a whole consists of three parts. $L_{vit}$ is used to train five ViT models for better prediction accuracy. $L_{lmc}$ is used to increase the difficulty of discrimination and make the network have better generalization ability. The purpose of $L_{fusion}$ is to consider local and global output results as a whole, so that the network as a whole does not depend too much attention on the result of each single network.

The input whole face is represented as $X_0$, and the four faces are divided into $X_{1,2,3,4}$ respectively. The input picture set is $X=\{X_0, X_1, X_2,X_3,X_4\}$. The real label of the picture is Y, 0 represents the real picture, 1 represents the reenactment picture, and the two categories are represented by C. $f_C (X_i)$is the likelihood that each network will predict true or false for input $X_i$. The cross entropy of the results of each VIT is calculated so that the network can learn the details of the forged image.
\begin{equation}
	\label{denoise}
	L_{ViT}=-\sum_{i=0}^{4}\sum_{c=0}^{1}Y_c\log f_c(X_i)
\end{equation}
We refer to the loss LMC in Cosface\cite{wang2018cosface} and modify the embedding of the whole face image extracted from the network according to image quality block. Using $e_i $ as representation of the image. $Y_i $ is the real classification of $e_i $ and $W_j$ is the category weight of j class. $\theta_j$ is the Angle between $e_i$ and $W_j$. Where s and m are superparameters.
\begin{equation}
	\label{denoise}
	W=\frac{W^*}{||W^*||},e=\frac{e^*}{||e^*||},
	\cos(\theta_j,i)=W_j^Te_i
\end{equation}
The $L_{lmc}$ used cosine margin to restrict the category of the current sample to the same category after subtracting m. After the information of picture quality is added, the network can dynamically set the size of constraints on pictures of different quality, and the network needs to learn more stricter criteria for distinguishing images of higher quality.
\begin{equation}
	\label{denoise}
	L_{lmc}=\sum_i-\log\frac{e^{s(\cos(\theta_{y_i},i)-m)}}{e^{s(\cos(\theta_{y_i},i)-m)}+\sum_{j\neq y_i}e^{s\cos(\theta_j,i)}}
\end{equation}

 $L_{fusion}$ refers to the cross entropy of the result of linear at the last layer in fusion attention block and Y, which is used to constrain the network to refer to the results of all ViT outputs. The final loss is, where the weight of $ L_{lmc}$ is represented by $\lambda$.
 \begin{equation}
 	\label{denoise}
 	Loss=\lambda L_{lmc}+L_{ViT}+L_{fusion}
 \end{equation}

\section{Experiment}

\subsection{Dataset}
Our training and testing were based on the FF++ dataset, which is a publicly available forgery video dataset widely used in the field of face reenactment. It contains 1000 real videos collected from the YouTube. Each video is guaranteed to be at least 300 frames, and the corresponding forged video is produced using Face2Face and other forge methods. For each video, the dataset also provides the masks corresponding to forged videos. Each video provides three compression versions according to H.264 codec, which are uncompressed (Raw), light compression(C23) and heavy compression(C40) by compression parameters.

The FF++ dataset provides the sort of training set, verification set and testing set. We randomly sampled 10 images within the number of video frames for each original video and corresponding forged video, and created data sets at the corresponding compression rate according to the compression rate.

\subsection{Details}
Our proposed model was implemented using the PyTorch framework and chose SGD as our optimizer with the default parameter. We tried to use AdamW as the optimizer, but it didn't perform well. All model trained 20 epochs at batch size 8 to achieve the best performance. The classifier is a ResNet-18 network, which pretrained on The ImageNet dataset. This classifier was also trained for 20 epochs, and the training parameters with the highest accuracy rate of 94\% were selected. All ViT models use The ImageNet-21K pretrained parameters. When the parameter in the loss function is 0.1, the whole model can achieves the best effect.

\section{Results and Observations}
We conducted a lot of experiments and compared our method with existing SOTA(state-of-the-art) methods. Kumar\cite{kumar2020detecting} proposed a multi-stream network model, using 5 ResNet-18 models, to detect and classify the whole face and partial face divided into four squares respectively. The experimental results show that the results of this method are better than many previous representative methods, so we reset the model according to the settings of the paper, and test the dataset we processed, and take the experimental results of this model as our main comparison object. At the same time, we also carried out replacement experiments with several models of different architectures to verify the advantages of our proposed model. In the case of retaining the original paper model architecture, backbone is replaced by five ViTs, one ViT with four ResNet-18 respectively.

\begin{table}[]
	\centering
	\caption{Accuracy (\%) of different models on the FaceForensics++ dataset with the same compression rate of train and test sets.
	}
	\label{tab:res1}
	\resizebox{\linewidth}{!}{%
		\begin{tabular}{|l|c|c|c|}
			\hline
			\rowcolor[HTML]{FFFFFF} 
			{\color[HTML]{333333} \textbf{Model}} & {\color[HTML]{333333} \textbf{Raw/Raw}} & {\color[HTML]{333333} \textbf{C23/C23}} & {\color[HTML]{333333} \textbf{C40/C40}} \\ \hline
			Multi-Resnet18\cite{kumar2020detecting}      & 99.74          & \textbf{99.49} & 87.17          \\ \hline
			Multi-ViT           & \textbf{99.96} & 99.32          & 88.39          \\ \hline
			ViT                 & 99.07          & 97.38          & 86.06          \\ \hline
			1V4R & 99.75          & 98.46          & 85.35          \\ \hline
			GGViT-base(Ours)           & 99.46          & 97.99          & 87.46          \\ \hline
			GGViT+IQB(Ours)            & 99.46          & 97.20          & 87.10          \\ \hline
			GGViT+FAB(Ours)            & 99.21          & 99.07          & 87.82          \\ \hline
			GGViT+IQB+FAB(Ours)        & 99.11          & 99.31          & \textbf{89.04} \\ \hline
		\end{tabular}%
	}
\end{table}

Table~\ref{tab:res1} shows the accuracy of the dataset at three different compression rates. Raw/Raw indicates that the train and test dataset of this experiment result are uncompressed data. It can be seen from the table that with the increase of image compression rate, the accuracy rate of network for image authenticity discrimination decreases. Multi-Resnet18 and Multi-ViT achieve the best accuracy at the two compression rates of Raw and C23 respectively. Our proposed model achieves the best results at C40 and also has excellent performance at the other two compression rates.
\begin{table*}[]
	\centering
	\caption{The accuracy of different models on the FaceForensics++ dataset with different compression rate of train and test sets}
	\label{tab:res2}
	\resizebox{\linewidth}{!}{%
		\begin{tabular}{|l|c|c|c|c|c|c|}
			\hline
			\rowcolor[HTML]{FFFFFF} 
			{\color[HTML]{333333} \textbf{Model}} &
			{\color[HTML]{333333} \textbf{Raw/C23}} &
			{\color[HTML]{333333} \textbf{Raw/C40}} &
			{\color[HTML]{333333} \textbf{C23/Raw}} &
			{\color[HTML]{333333} \textbf{C23/C40}} &
			{\color[HTML]{333333} \textbf{C40/Raw}} &
			{\color[HTML]{333333} \textbf{C40/C23}} \\ 
			\hline
			Multi-Resnet18\cite{kumar2020detecting}      & 66.81          & 50.27          & \textbf{99.49} & 55.39          & 92.73          & 90.93         
			\\ \hline
			Multi-ViT           & 70.57          & 53.10          & 98.57          & 62.38          & 94.38          & 92.61          \\ \hline
			1V4R & 65.84          & 50.19          & 98.25          & 58.19          & 88.36          & 86.52          
			\\ \hline
			GGViT-base(Ours)           & 83.41          & 58.44          & 99.46          & 72.84          & 91.05          & 90.07          \\ \hline
			GGViT+IQB(Ours)            & 89.18          & 60.19          & 99.14          & \textbf{73.63} & 94.24          & 92.29          \\ \hline
			GGViT+FAB(Ours)            & 87.31          & 61.84          & 99.14          & 70.33          & 92.55          & 90.36          \\ \hline
			GGViT+IQB+FAB(Ours)        & \textbf{91.15} & \textbf{65.35} & 99.07          & 65.53          & \textbf{94.46} & \textbf{93.51} \\ \hline
			
		\end{tabular}%
	}
\end{table*}

According to the results from Table~\ref{tab:res2}, ViT can greatly improve the generalization ability of the model on datasets with different compression rates. Raw/C23 indicates that the training dataset for this experiment is Raw and the test dataset is C23. In training sets and test sets of different compression rates, the multi-ViT has 1.71\%-3.76\% percentage improvement over the multi-ResNet18, especially the C23/C40 with $6.99\%$ improvement, except the C23/Raw combination multi-ResNet18 has a slight lead.

Based on the multi-ResNet18 model, 1V4R replaces the backbone of the stream network for detecting the whole face with Vit to keep the rest of the network structure unchanged. The goal is to use ResNet to reduce network computing while allowing ViT to use global information to guide individual stream networks. However, the effect is not ideal, only C23/C40 has a $2.8\%$ improvement compared with multi-Resnet18. We speculate that the difference between the two models makes the global information cannot assist each stream well.

The GGViT-base model we proposed has been greatly improved by 16.67$\%$, $8.17\%$, and $17.45\%$ on Raw/C23,Raw/C40 and C23/C40, respectively. Compared with the multi-ViT model, it has also increased by $12.84\%$,$ 5.34\%$ and $10.46\%$ respectively. which shows that adding the ViT output result of the entire face to the segmented face part picture can provide global information to the stream network responsible for judging the face part image, and help the local stream network make judgments. When training on the dataset with a lower compression rate and testing on the dataset with a higher compression rate, the traces of image forgery are destroyed due to compression, and it is difficult for the local stream network to learn the basis for the judgment of forgery. Global information can effectively help improve the discriminating ability of the local stream network, thereby improving the generalization ability of the model.

Comparing the use of the IQB(image quality block) module under the two network architectures, it can be found that because high-quality pictures have more detailed information, there is a big gap between different types of images, so the image quality block makes the network more stringent for high-quality images discrimination criteria, and as the quality of the images decreases, the difference between the real images and the fake images becomes less and less. The image quality block's improvement to the network is the most obvious on Raw/C23, an increase of 5.77$\%$.

Forged images information will be changed after compression, this makes the network pays attention to different areas of images of different compression rates. FAB(fusion attention block) aims to integrate the predictions of multi-stream networks, allowing the network to pay more attention to the local areas that can contain more traces of forgery under the current image quality. The proportion of the results of each stream network in the tensor we use for our final determination is shown in Table~\ref{tab:res3}. Where X0 and X1-X4 respectively represent the output results of five stream networks, namely, the whole face, upper left part, upper right part, lower left part and lower right part respectively.

According to the data results of the training set and test set in the table are both raw, the network mainly relies on the data information on X3 and X4 stream networks to judge the uncompressed pictures, and the output results in these two networks account for $72.3\%$ of the final judge tensor. As the image is compressed, the proportion of X3 and X4 also decreases, accounting for $64.68\%$ on C23/C23 and $59.42\%$ on C40/C40. The proportions of X1 and X2 are $18.56\%$, $29.36\%$ and $32.36\%$ respectively, showing an increasing trend as a whole. This is particularly evident in the increasing of X1 and the decreasing of X3. Combined with the overall experimental results, the networks trained in C23 dataset pay more attention to X1 and X3, while the networks trained in C40 dataset pay more attention to X1 and X4.

As the FAB pays more attention to the stream network output results that can help distinguish from the current compression rate, in order to obtain the best classification results for the network trained on C23 data, X3 accounts for more than $50\%$ of the most fusion tensor, and X4 accounts for only $7.36\%$. This also leads to the poor performance of the network trained on C23 on C40, because C40 mainly depends on the network output of X4. Even so, compared with multi-ResNet18, our proposed method still achieves a 10.14\% improvement on C23/C40.

Our approach combines the advantages of multi-stream networks and ViT. Each stream network not only improves the accuracy of classification, but also improves its generalization ability through global information. This also indicates that the previous method does not make good use of the information of the whole picture to guide each stream network. Based on this, our method also designs corresponding blocks for image quality and final prediction results, so it can have a good performance.

\begin{table}[]
	\centering
	\caption{The percentage(\%) of the output results of each stream network in the final discriminant feature}
	\label{tab:res3}
	\resizebox{\linewidth}{!}{%
		\begin{tabular}{|c|c|c|c|c|c|}
			\hline
			\textbf{Train/Test} & \textbf{X0} & \textbf{X1} & \textbf{X2} & \textbf{X3} & \textbf{X4} \\ \hline
			Raw/Raw             & 9.14        & 9.06        & 9.5         & 31.74       & 40.56       \\ \hline
			Raw/C23             & 10.9        & 15.86       & 13.06       & 28.8        & 31.38       \\ \hline
			Raw/C40             & 15.38       & 43.96       & 23.16       & 8.84        & 8.66        \\ \hline
			C23/Raw             & 5.24        & 12.18       & 10.26       & 64.42       & 7.9         \\ \hline
			C23/C23             & 5.96        & 17.44       & 11.92       & 57.32       & 7.36        \\ \hline
			C23/C40             & 12.44       & 38.64       & 17.46       & 27.58       & 3.88        \\ \hline
			C40/Raw             & 10.04       & 13.66       & 9.22        & 25.62       & 41.46       \\ \hline
			C40/C23             & 10.04       & 15.64       & 8.46        & 20.52       & 45.34       \\ \hline
			C40/C40             & 8.22        & 21.92       & 10.44       & 14.6        & 44.82       \\ \hline
		\end{tabular}%
	}
\end{table}

\section{Conclusion}
In this paper, we proposed our cross domain discrimination solution for forged face video with different compression rates in social media. Compared with other networks with single compression rate, our method achieves SOTA on FF++ dataset And it has a significant improvement on datasets with different compression rates. The ablation experiments show that the proposed modules are helpful for the network to deal with the classification of images under different compression conditions. We also explored the change of network attention to images with different compression rates, and explained the experimental results.

Our future goal is to provide more guidance for the areas to be concerned through image quality on cross domain issues, reduce the calculation scale of the model, and make the use scenario of the model more suitable for a variety of social media terminals.





%
\bibliographystyle{IEEEtran}
\bibliography{ref}

\end{document}